\documentclass[letterpaper, 10 pt, conference]{ieeeconf}  

\IEEEoverridecommandlockouts                              

\usepackage{graphicx} 
\usepackage{epsfig} 
\usepackage{mathptmx} 
\usepackage{times} 
\usepackage{amsmath} 
\usepackage{amssymb}  
\usepackage{bm}
\usepackage{subfig}
\usepackage{url}
\usepackage{color}
\usepackage[ruled]{algorithm2e}
\DeclareMathOperator*{\argmin}{argmin}
\title{\LARGE \bf
	MIS-SLAM: Real-time Large Scale Dense Deformable SLAM System in Minimal Invasive Surgery Based on Heterogeneous Computing
}

\author{Jingwei Song$^{}$, Jun Wang$^{}$, Liang Zhao$^{}$, Shoudong Huang$^{}$ and Gamini Dissanayake$^{}$
	\thanks{$^{}$All the authors are from Centre for Autonomous Systems, University of Technology, Sydney, P.O. Box 123, Broadway, NSW 2007, Australia}%
	\thanks{Email: jingwei.song@student.uts.edu.au, wangjun@radi.ac.cn, \{Liang.Zhao; Shoudong.Huang;
		Gamini.Dissanayake\}@uts.edu.au} 
}

\begin{document}

	\maketitle
	\thispagestyle{empty}
	\pagestyle{empty}

	\begin{abstract}
		Real-time simultaneously localization and dense mapping is very helpful for providing Virtual Reality and Augmented Reality for surgeons or even surgical robots. In this paper, we propose MIS-SLAM: a complete real-time large scale dense deformable SLAM system with stereoscope in Minimal Invasive Surgery based on heterogeneous computing by making full use of CPU and GPU. Idled CPU is used to perform ORB-SLAM for providing robust global pose. Strategies are taken to integrate modules from CPU and GPU. We solved the key problem raised in previous work, that is, fast movement of scope and blurry images make the scope tracking fail. Benefiting from improved localization, MIS-SLAM can achieve large scale scope localizing and dense mapping in real-time. It transforms and deforms current model and incrementally fuses new observation while keeping vivid texture. In-vivo experiments conducted on publicly available datasets presented in the form of videos demonstrate the feasibility and practicality of MIS-SLAM for potential clinical purpose.   
		
	\end{abstract}

	\section{INTRODUCTION}
	
	Comparing with open surgery, Minimally Invasive Surgery (MIS) brings shortcomings such as lack of field of view, poor localization of scope and fewer surrounding information. Limited by these, surgeons are required to perform the intra-operations in narrow space with elongated tools and without direct 3D vision \cite{mountney2009dynamic}. To overcome these problems, surgeons spend large amount of time training to be familiar with doing operations under scopes. \par 
	
	SLAM, simultaneous localization and mapping, is a technique applied in robotics for pose estimation and environment mapping. Efforts have been devoted to exploit the feasibility of applying SLAM to localize the scope and reconstruct a sparse or even dense soft-tissue surface. \cite{grasa2011ekf} and \cite{lin2013simultaneous} adopt conventional feature based SLAM, these are extended Kalman filter (EKF) SLAM and Parallel Tracking and Mapping (PTAM). They improved EKF and PTAM by using threshold strategies to separate rigid and non-rigid feature points. Mahmoud et al. \cite{mahmoud2016orbslam}\cite{mahmoud2017slam}\cite{turan2017non} exploit and tune a complete and widely used large scale SLAM system named ORB-SLAM \cite{mur2015orb}. They analyze and proved that ORB-SLAM is also suitable for scope localization in MIS. In \cite{mahmoud2017slam}, a quasi-dense map can be estimated off-line based on pose imported from ORB-SLAM. Feature points for localization are widely used for sparse mapping. Contrary to feature based SLAM, Du et al. \cite{du2015robust} adopt dense matching SLAM which employed a special optical flow namely Deformable Lucas-Kanade for tracking tissue surface. Aside from SLAM, other approaches also contribute greatly to enable augmented reality (AR) and virtual reality (VR) in MIS. \cite{stoyanov2012stereoscopic} proposes an approach to recover 3D geometry from stereo images. A structure from motion pipeline \cite{lin2015video} is proposed for partial 3D surgical scene reconstruction and localization.  \cite{haouchine2015monocular} and \cite{malti2011template} extract whole tissue surface from stereo or monocular images. All these approaches contribute greatly to MIS. However, they still don't provide a real-time complete and robust solution for localizing scope while reconstructing dense deformable soft-tissue surfaces. All the SLAM techniques mentioned above focus on monocular scope and didn't solve the problem of missing scale, thus making localization not very practical.\par 
	
	To broaden surgeons' field of view, 3D laparoscopy, or binoculars is applied to generate 2 images from different viewing point so that a 3D geometry based on parallax is created in surgeons' mind for better understanding of the environment. Recently, similar stereo vision is adopted by some AR devices for enhancing MIS procedures. Therefore, it will be very helpful if stereo vision related approaches in computer vision community could be integrated, extended and improved to recover deformable shape in real-time while estimating the pose of the camera. In our previous work, we proposed a dynamic reconstruction of deformable soft-tissue with stereo scope \cite{song2018dynamic}. A warping field based on the
	embedded deformation nodes approach  is introduced with 3D dense shapes
	recovered from sequences pairs of stereo images. With the help of general-purpose
	computing on graphics processing units (GPGPU), all the processes are achieved in real-time. Mentioned in \cite{song2018dynamic}, the first and most important challenge to the pipeline is the fast movement of scope. Fast movement not only makes visual odometry unstable but also causes blurry images which make registrations even worse. This issue has also been reported in \cite{grasa2011ekf} and \cite{lin2013simultaneous}. \par

	\begin{figure*}[]
		\centering
		\includegraphics[width=1\textwidth]{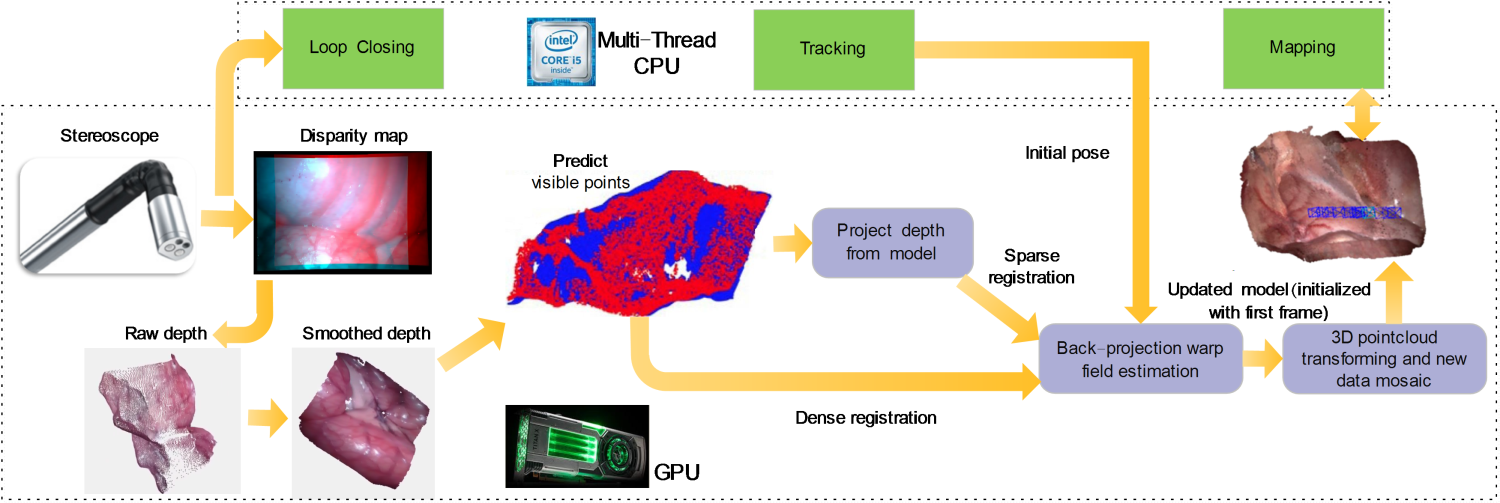}
		\caption{The framework of MIS-SLAM. CPU is responsible for ORB-SLAM, uploading features, global pose and organize a visualization module. GPU processes depth estimation, registration, fusion and visualization.}
		\label{fig:pipeline}
	\end{figure*}

	Inspired by the research done by Mahmoud et al. \cite{mahmoud2016orbslam} \cite{mahmoud2017slam} \cite{turan2017non} which demonstrate the robustness of camera pose estimation from ORB-SLAM, we figure out ORB-SLAM is suitable to be improved and coupled with dense deformable SLAM. In this paper, we propose MIS-SLAM which builds on our preliminary work \cite{song2018dynamic} with the following major improvements: (1) We proposed a heterogeneous  framework to make full use of both GPU (dense deformable SLAM) and CPU (ORB-SLAM) to recover the dense deformed 3D structure of the soft-tissues in MIS scenario. Computational power of CPU is fully exploited to run an improved ORB-SLAM to provide complementary information to GPU modules. (2) Modules from GPU and CPU are deeply integrated to boost performance and enhance the efficiency. Sparse ORB features as well as global pose are uploaded to GPU. (3) We upgrade point cloud map and fusion management strategy to enhance large scale soft-tissue reconstructing. Comparing with truncated signed distance function (TSDF) widely used in computer vision community \cite{newcombe2015dynamicfusion}, \cite{innmann2016volumedeform} and \cite{dou2016fusion4d}, point cloud management in MIS-SLAM  notably reduce memory as well as boost the performance. (4) Real-time visualization is achieved on GPU end. MIS-SLAM can process large scale surface reconstruction in just one desktop in real-time. We suggest readers to view the associate video for the reconstruction process and
	to fully appreciate the live capabilities of the system.
	\par
	

	\section{Technical Details}
	
	\subsection{Overview of MIS-SLAM}
	
	The framework of MIS-SLAM includes improved ORB-SLAM (CPU), warping field and global pose estimation (GPU), new data fusion (GPU), data transforming from CPU to GPU and GPU based visualization. ORB-SLAM is first launched on CPU; ORB features and global pose are uploaded from CPU to GPU global memory. After receiving initial global pose from CPU, GPU modules first initialize the model with first estimated depth. Each time when new observation is acquired, the matched ORB features are upload to GPU. Potential visible points are extracted from model and projected into 2D depth images. System performs a registration process to estimate optimum rigid transformation as well as non-rigid warping field. Live model is then deformed to current state according to this transformation and fused with new observation. We make full use of the feature called `Graphic Interoperability` in compute unified device architecture (CUDA) to directly visualize model from GPU side. Fig. \ref{fig:pipeline} demonstrates the pipeline of these processes.\par

	Realizing the point cloud generated from stereo images are much less reliable than depth perception sensors, we modify and update previous point cloud approach with more properties. Each point stores five domains: coordinate $\bm{v}_i$, weight $\omega(\bm{v}_i)$, color $\bm{C_i}$, time stamp $\bm{t_i}$ and a boolean variable stability $\bm{S_i}$. Comparing with conventional TSDF based approach, point cloud management enables processing larger scale because empty space is not computed, no predefined space is calculated and marching cube process in grid-based approach can be avoided.\par 
	

	\subsection{Depth estimation from stereo images}
\begin{figure}[h]	
	 	\centering
	 	\subfloat[Original depth]{	
	 		\begin{minipage}[htpb]{0.22\textwidth}	
	 			\centering
	 			\includegraphics[width=1\linewidth]{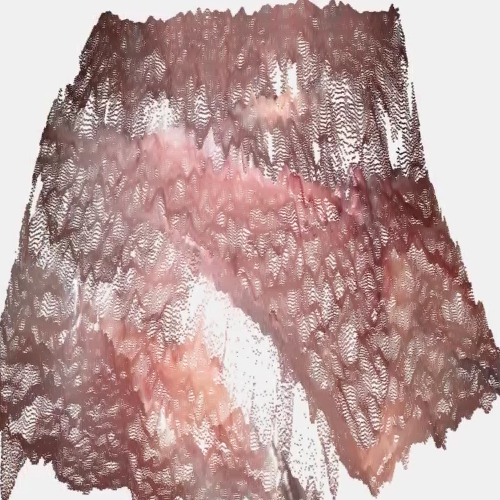}			
	 		\end{minipage}				
	 	}
	 	\subfloat[Fused model]{
	 		\centering
	 		\begin{minipage}[htpb]{0.22\textwidth}
	 			\centering
	 			\includegraphics[width=1\linewidth]{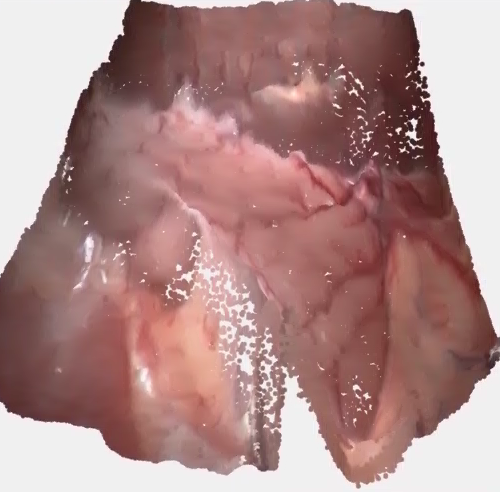}
	 		\end{minipage}
	 	}\/
	 	\caption{Examples of depth and fused model.}
	 	\label{fig:depthandmodel}
	 \end{figure}

	Efficient Large-scale Stereo (ELAS) \cite{billings2012system} is adopted as the depth estimation method. ELAS has been widely proved to achieve good result in surgical vision \cite{zhang2017autonomous}. Fig. \ref{fig:depthandmodel} is the example of original depth and fused model. \par 
	 \captionsetup[subfigure]{labelformat=empty}
	 \vspace{0pt} 
	
	\subsection{Sparse key correspondences and camera pose estimation}
	\label{SparseKeyPointsSec}
	With regard to the main issue in our previous work \cite{song2018dynamic}, we figure out the inaccuracy of global scope pose leads to instability of the pipeline. In order to improve the robustness of the system, we make full use of idled CPU to implement robust large scale SLAM system and couple CPU and GPU strategically. This was motivated by the fact that several successful large scale SLAM system has been deliberately developed, trained and tuned for enhancing robustness in multi-levels including architectures, algorithms, thresholds, tricks and codes. Importantly, due to computational power limit and bandwidth, efforts should be devoted to exploring how to make full use of this integration. ORB-SLAM module provides the ORB features which are fully exploited on GPU. This strategy save computational powers on GPU: (1) Dense Speeded Up Robust Feature (SURF) extraction and match step in our original approach \cite{song2018dynamic} is therefore substituted. (2) Visual Odometry (VO) based global pose initialization is abandoned. (3) Random sample consensus (RANSAC) is also not needed anymore.

	\subsection{Deformation}
	
	The basic idea of deformation description is weighted average of locally rigid rotation and transformation defined by deformation nodes, which are sparsely and evenly scattered in space. Each node j comes with affine matrix $\bm{A_j}$ $\in\mathbb{R}^{3\times3}$ and a translation vector in $\bm{t_j}$ $\in\mathbb{R}^3$. Each source point is transformed to its target position by several nearest embedded deformation (ED) nodes. ED node $j$ is described by a position $\bm{g_j}$ $\in\mathbb{R}^3$, a corresponding quasi rotation (affine) matrix  and a translation vector. For any given vertex $\bm{p}$, the new position $\bm{\tilde{p}}$ is defined by the ED node $j$ as:\par
	\begin{equation}
	\bm{\tilde{\bm{v}}_i}=\bm{R}_i{\sum_{j=1}^k w_j(\bm{\bm{v}_i})[\bm{A_j}(\bm{\bm{v}_i}-\bm{g_j})+\bm{g_j}+\bm{t_j}]}+\bm{T}_i
	\label{TransformationFomulation}
	\end{equation}
	where $k$ denotes number of neighboring node. $w_j(\bm{\bm{v}_i})$ is quantified weight for transforming $\bm{\bm{v}_i}$ exerted by each related ED node. $\bm{R}_i$ and $\bm{T}_i$ denote rigid rotation and translation. We confine the number of nearest nodes by defining the weight in Eq. \ref{eq_weight}. Deformation of each point in the space is limited locally by setting the weight as:
	\begin{equation}
	\label{eq_weight}
	w_j(\bm{\bm{v}_i})=(1-||\bm{\bm{v}_i}-\bm{g_j}||/d_{max})
	\end{equation}
	where $d_{max}$ is the maximum distance of the vertex to $k + 1$ nearest ED node.

	\subsection{Energy function}
	The objective function formulated is composed of six terms: Rotation constraint, Regularization, the point-to-plane distances between the visible points and the target scan, sparse key points correspondence and global pose as:\par
	\begin{equation}
	\begin{split}
	\argmin\limits_{\bm{R}_i,\bm{T}_i,\bm{\bm{A_1}},\bm{\bm{t_1}}...\bm{A_m},\bm{t_m}} &w_{rot}E_{rot}+w_{reg} E_{reg}+w_{data} E_{data}+w_{corr} E_{corr}\\
	&+w_{r} E_{r}+w_{p} E_{p}
	\end{split}
	\label{energyfunction}
	\end{equation}
	where $m$ is the number of ED nodes.  \par
	To prevent the optimization converging to an unreasonable deformation, we follow the method proposed in \cite{sumner2007embedded} which constrains the model with Rotation and Regularization.  \par
	\textbf{Rotation}. $E_{rot}$ sums the rotation error of all the matrix in the following form:
	\begin{equation}
	E_{rot}=\sum_{j=1}^m Rot(\bm{A_j})
	\end{equation}
	\begin{equation}
	\begin{aligned}
	Rot(\bm{A})=(\bm{c_1}\cdot\bm{c_2})^2+(\bm{c_1}\cdot\bm{c_3})^2+(\bm{c_2}\cdot\bm{c_3})^2+\\
	(\bm{c_1}\cdot\bm{c_1}-1)^2+(\bm{c_2}\cdot\bm{c_2}-1)^2+(\bm{c_3}\cdot\bm{c_3}-1)^2
	\end{aligned}
	\end{equation}
	where $\bm{c_1}$, $\bm{c_2}$ and $\bm{c_3}$ are the column vectors of the matrix $\bm{A}$.
	
	\textbf{Regularization}. This term is to prevent divergence of the neighboring nodes exerts on the overlapping space. For details, please refer to \cite{song2018dynamic}. \par
	\begin{equation}
	E_{reg}=\sum_{j=1}^m\sum_{k\in{\mathbb{N}(j)}} \alpha_{jk}||\bm{A_j}(\bm{\bm{g_k}}-\bm{g_j})+\bm{g_j}+\bm{t_j}-(\bm{\bm{g_k}}+\bm{t_k})||^2
	\end{equation}
	where $\alpha_{jk}$ is the weight calculated by the Euclidean distance of the two ED nodes. We follow \cite{sumner2007embedded} by uniformly setting $\alpha_{jk}$ to 1. $\mathbb{N}(j)$ is the set of all neighboring nodes to the node $j$. \par
	\noindent\textbf{Data Term}. 
	We follow Algorithm \ref{Fusionstep} to find registrations of model points and calculate point to plane distance of all the registered points. Thresholds are used to extract visible points $\bm{v}_i ~(i\in\{1,...,N\})$. \par
	\begin{equation}
	||\bm{\bm{v}}_i-\Gamma(P(\bm{v}_i))|| < \epsilon_d,  \quad
	\bm{\textbf{n}}(\bm{v}_i) \cdot \bm{\textbf{n}}(P(\bm{v}_i)) < \epsilon_n
	\end{equation}
	where $\epsilon_d$ and $\epsilon_n$ are thresholds. \par 
	We adopt back-projection approach as a model-to-scan registration strategy that penalizes misalignment of the predicted visible points $\bm{v}_i ~(i\in\{1,...,N\})$ and current depth scan $\mathbb{D}$. Data Term is sum of point-to-plane errors in the form of: \par
	\begin{equation}
	E_{data}=\sum_{i=1}^N (\bm{\textbf{n}}(P(\tilde{\bm{v}_i}))^T(\bm{\tilde{\bm{v}}_i}-\Gamma(P(\tilde{\bm{v}_i})）))^2
	\end{equation}
	where $\Gamma(\bm{v}) = \Pi(P(\bm{v}))$ and $\tilde{\bm{\textbf{n}}}(u)$ is the corresponding normal to the pixel $u$ in the depth $\mathbb{D}(u)$. $\bm{\tilde{\bm{v}}_i}$ is the transformed position of point $\bm{v}_i$. $P(\bm{v})$ is the projective ($\mathbb{R}^3 \rightarrow \mathbb{R}^2$) function for projecting visible points to depth image. $\Pi(u)$ is the back-projection function for lifting the corresponding pixel $P(\bm{v})$ back into 3D space ($\mathbb{R}^2 \rightarrow \mathbb{R}^3$). \par

	As described in \cite{dou2016fusion4d}, back-projection and point-to-plane strategies make full use of the input depth image so that the Jacobians can be calculated in 2D which leads to fast convergence and robustness to outliers. \par
	\noindent\textbf{Correspondence}. It is measured by the Euclidean distance between pair-wise sparse key points generated from dense SURF described in Section  \ref{SparseKeyPointsSec} in the following form:
	\begin{equation}
	E_{corr}=||\bm{\tilde{\bm{V}}_i-\bm{V}_i||}
	\end{equation}
	where $\tilde{\bm{V}}_i$ and $\bm{V}_i$ are the new and old position of ORB features. \par 
	
	\noindent\textbf{Global Pose}. It is measured by the variations of rotation and transformation. The global pose is defined by pose relative to the first frame. We use Euclidean distance and Euler angles to define the variation between optimized global pose and original global pose generated by ORB-SLAM. It is presented in the following form:
	\begin{equation}
	\begin{split}
	&E_{r}=||\bm{\tilde{\bm{R}}_i-\bm{R}_i||}\\
	&E_{p}=||\bm{\tilde{\bm{P}}_i-\bm{P}_i||}
	\end{split}
	\end{equation}
	where $\tilde{\bm{R}}_i$ and $\bm{R}_i$ are the new and old orientation of the scope; $\tilde{\bm{P}}_i$ and $\bm{P}_i$ are the new and old position of the scope.\par 
	
	\subsection{Optimization}
	We follow our previous optimization step using Levenberg-Marquardt (LM) to solve the nonlinear optimization problem. The efficiency is almost the same as only 6 more variables (Global orientation and translation) are added.

	\subsection{Model update with new observation}
	For each point ($\bm{v}_i$, $\omega(\bm{v}_i)$, $\bm{C_i}$, $\bm{t_i}$, $\bm{S_i}$) in one fusion step, all the properties of each point is updated. Moreover, inspired by this approach \cite{keller2013real}, we modify map management module by adding a time stamp property and introduce stability control. The general algorithm is described in Algorithm \ref{Whole deform and fusion Process}.\par
	
	After estimating the optimum warping field, we transform all the points to their best positions and predict visible points again (Algorithm \ref{Fusionstep}). For each updated point, truncated signed distance weight (TSDW) is assigned to each pixel of new depth: 
	\begin{equation}
	\omega(\bm{p}_n) =
	\begin{cases}
	d_{min}(\bm{p}_n) / (0.5 * \epsilon) &\mbox{if $abs(\bm{\bm{\tilde{p}}_i}|_z - \mathbb{D}(P(\bm{p}_n)))<\tau$}\\
	0 &\mbox{otherwise}
	\end{cases}
	\end{equation}
	
	\setlength{\parindent}{0pt}where $d_{min}(\bm{p})$ is the minimum distance of point $\bm{p}$ to its corresponding nodes and $\tau$ is the grid size of nodes. We discard the $z$ directional difference if it is too large as this is probably due to inaccurate warp field estimation.\par 
	\begin{algorithm}
		\KwIn{Point cloud state in last frame $\bm{p}_{n-1}$; Depth map in current state $\mathbb{D}_{n+1}$}
		\KwOut{Point cloud state in current state $\bm{p}_n$}
		\While{loop over each pixel within $Reg$}
		{
			\If{$\mathbb{D}(P(\tilde{\bm{p}_i})) is not empty$}
			{
				\If{
					(1.$abs(\bm{\bm{\tilde{v}}_i}|_z - \mathbb{D}(P(\bm{p}_n)))<\tau$\\
					2.$angle(\bm{\bm{\tilde{n}}} - \mathbb{N}(N(\bm{n}_n)))<\delta$)}
				{
					Find registration
				}
			}
		}
		
		\caption{Point to depth registration}
		\label{Fusionstep}
	\end{algorithm} 
	\begin{algorithm}
		\KwIn{Point cloud state in last frame $\bm{p}_{n-1}$; Depth map in current state $\mathbb{D}_{n+1}$}
		\KwOut{Point cloud state in current state $\bm{p}_n$}
		Step 1:\\
		Register point cloud with current depth. Fuse each point with depth and lift unregistered depth back into space.\\
		Step 2:\\
		\For{$i$=1;$i < $Number of point cloud;$i++$}
		{
			Deform $\bm{p}_{n-1}$ according to estimated warping field\\
			Project deformed point cloud to depth map to build a point to pixel registration map $Reg$. \\
		}
		Step 3:\\
		Implement Algorithm \ref{Fusionstep}\\
		\While{loop over each pixel within $Reg$}
		{
			\eIf{Pixel is registrated}
			{
				Fuse points following formulation \ref{fusion1},  \ref{fusion2}, \ref{fusion3} and \ref{fusion4}.	
				Organize points to Group1\\ 
			}
			{
				Lift new pixel into 3D space
				Organize points to Group 2\\
			}
		}
		
		Step 4:\\
		\While{loop over each point $p_k$ in  Group1}
		{
			\If{1.$p_k{time} < \tau_{time}$  \\
				2.$p_k{weight} < \tau_{weight}$}
			{
				Delete $p_k$
			}
		}
		Step 5:\\
		Merge Group 1 and Group 2 to new group.\\
		Regenerate node and corresponding rotation and translation. 
		
		\caption{Point cloud based fusion with depth images in MIS-SLAM}
		\label{Whole deform and fusion Process}
	\end{algorithm} 
	
	\begin{algorithm}
		\label{algo2}
		\KwIn{Point cloud state in last frame $\bm{p}_{n-1}$; Depth map in current state $\mathbb{D}_{n+1}$}
		\KwOut{Point cloud state in current state $\bm{p}_n$}
		Down sample point cloud according to grid size.\\
		\While{loop over each point $p_k$ in  Group1}
		{
			\eIf{$t_k < \tau_{time}$  
				and $\omega_k < \tau_{weight}$}
			{
				Delete $p_k$
			}
			{
				Stamp $p_k$: $t_k$ = current frame index\\
			}
		}
		\caption{Removing unwanted, noisy or redundant point cloud}
		\label{Removepoints}
	\end{algorithm} 
	We fuse current depth generated from model with new depth by:
	\begin{equation}
	\label{fusion1}
	\mathbb{D}_{n+1}(P(\bm{p}_n)) = \frac{\tilde{\bm{v}}_i|_z \omega(\bm{p}_{n-1})+\mathbb{D}_n(P(\bm{p}_n))}{\omega(\bm{p}_{n-1})+1}
	\end{equation}
	\begin{equation}
	\label{fusion2}
	\mathbb{C}_{n+1}(P(\bm{p}_n)) = \frac{\tilde{\bm{C}}_i \omega(\bm{p}_{n-1})+\mathbb{C}_n(P(\bm{p}_n))}{\omega(\bm{p}_{n-1})+1}
	\end{equation}
	\begin{equation}
	\label{fusion3}
	\mathbb{N}_{n+1}(N(\bm{n}_n)) = \frac{\tilde{\bm{n}}_i \omega(\bm{p}_{n-1})+\mathbb{N}_n(N(\bm{n}_n))}{\omega(\bm{p}_{n-1})+1}
	\end{equation}
	\begin{equation}
	\label{fusion4}
	\omega(\bm{p}_n)=min(\omega(\bm{p}_{n-1})+1 ,\omega_{max})
	\end{equation}
	where $\tilde{\bm{v}}_i|_z$ is the value of point $\bm{v}_i$ on the z direction and $\bm{p}$ is the projected pixel $P(\tilde{\bm{v}}_i)$ of the transformed points $\bm{v}_i$. $\mathbb{D}_{n+1}$, $\mathbb{C}_{n+1}$ and $\mathbb{N}_{n+1}$ are depth map, color map and normal maps respectively. Different from rigid transformation where uncertainty of all the points in 3D space are considered as equal, in the case of non-rigid fusion, if a point is further away to the nodes of warping field, we are less likely to believe the registered depth \cite{newcombe2015dynamicfusion}. Therefore, we practically measure this certainty by using the minimum distance from point to nodes and regularize it with half of the unified node distance. The upper bound of weight is set to 10. Algorithm \ref{Whole deform and fusion Process} and Eq. (\ref{fusion1})(\ref{fusion2})(\ref{fusion3})(\ref{fusion4}) show the details for point fusion. \par
	In order to filter noises, we apply the property `stability $\bm{S_i}$' to describe each point. Unstable points which appear in few frames and are not visible in recent frame are abandoned. Algorithm \ref{Removepoints} shows the strategy of point filtering.\par
	
	Our weighted points based method offers numbers of benefits: (1) Points based data management is free of extent limitation in traditional TSDF based approach. With our fusion based Algorithm  \ref{Removepoints}, fused geometry can still keep its shape smooth. (2) All the components in MIS-SLAM, i.e. visible points prediction, warping field estimation and model update are unified in points. Process like conversion between volume and mesh are not necessary anymore.
	
	\captionsetup[subfigure]{labelformat=empty}
	\vspace{0pt} 
	\begin{figure*}[h]	
		\label{resultmodel}
		\centering
		\subfloat[Frame 1]{	
			\begin{minipage}[htpb]{0.16\textwidth}	
				\centering
				\includegraphics[width=1\linewidth]{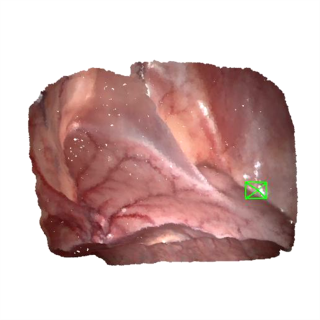}			
			\end{minipage}				
		}%
		\subfloat[Frame 100]{
			\centering
			\label{fig:scanmatching_b}
			\begin{minipage}[htpb]{0.16\textwidth}
				\centering
				\includegraphics[width=1\linewidth]{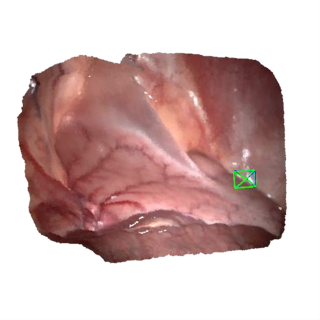}
			\end{minipage}
		}
		\subfloat[Frame 200]{
			\label{fig:scanmatching_c}
			\begin{minipage}[htpb]{0.16\textwidth}
				\centering
				\includegraphics[width=1\linewidth]{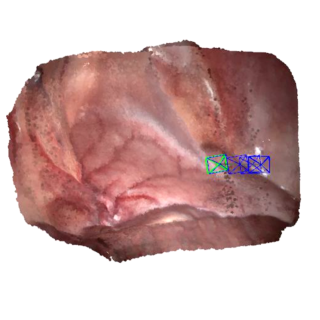}
			\end{minipage}
		}
		\subfloat[Frame 300]{
			\centering
			\label{fig:scanmatching_b}
			\begin{minipage}[htpb]{0.16\textwidth}
				\centering
				\includegraphics[width=1\linewidth]{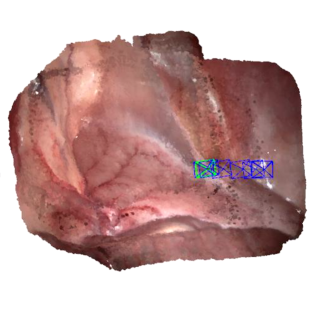}
			\end{minipage}
		}
		\subfloat[Frame 400]{
			\centering
			\label{fig:scanmatching_b}
			\begin{minipage}[htpb]{0.16\textwidth}
				\centering
				\includegraphics[width=1\linewidth]{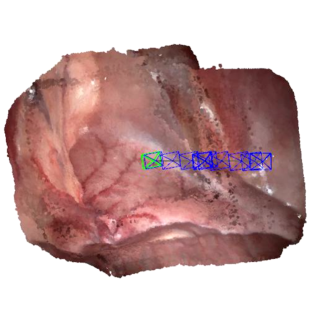}
			\end{minipage}
		}\/	
		\subfloat[Frame 500]{	
			\label{}
			\begin{minipage}[htpb]{0.16\textwidth}	
				\centering
				\includegraphics[width=1\linewidth]{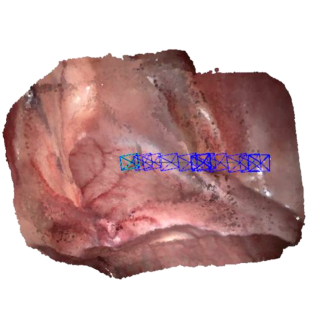}			
			\end{minipage}				
		}%
		\subfloat[Frame 600]{
			\centering
			\label{fig:scanmatching_b}
			\begin{minipage}[htpb]{0.16\textwidth}
				\centering
				\includegraphics[width=1\linewidth]{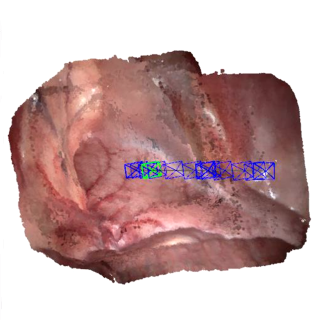}
			\end{minipage}
		}
		\subfloat[Frame 700]{
			\label{fig:scanmatching_c}
			\begin{minipage}[htpb]{0.16\textwidth}
				\centering
				\includegraphics[width=1\linewidth]{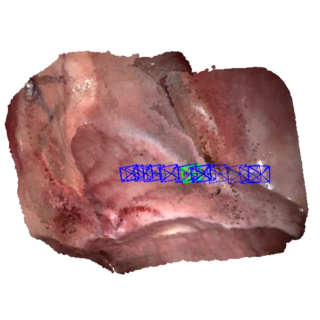}
			\end{minipage}
		}
		\subfloat[Frame 800]{
			\centering
			\label{fig:scanmatching_b}
			\begin{minipage}[htpb]{0.16\textwidth}
				\centering
				\includegraphics[width=1\linewidth]{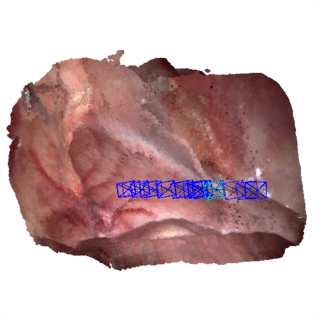}
			\end{minipage}
		}
		\subfloat[Frame 900]{
			\centering
			\label{fig:scanmatching_b}
			\begin{minipage}[htpb]{0.16\textwidth}
				\centering
				\includegraphics[width=1\linewidth]{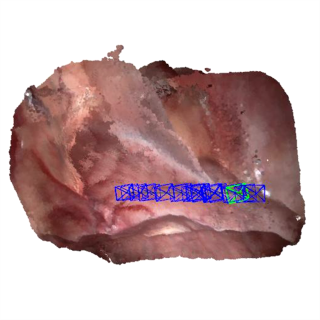}
			\end{minipage}
		}\/
		\subfloat[Frame 1]{
			\label{fig:scanmatching_a}
			\begin{minipage}[htpb]{0.16\textwidth}
				\centering
				\includegraphics[width=1\linewidth]{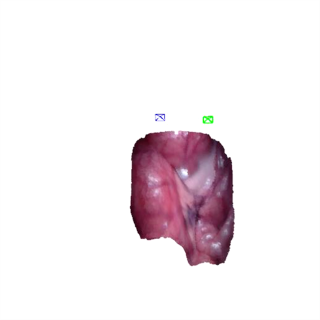}
			\end{minipage}
		}%
		\subfloat[Frame 200]{
			\centering
			\label{fig:scanmatching_b}
			\begin{minipage}[htpb]{0.16\textwidth}
				\centering
				\includegraphics[width=1\linewidth]{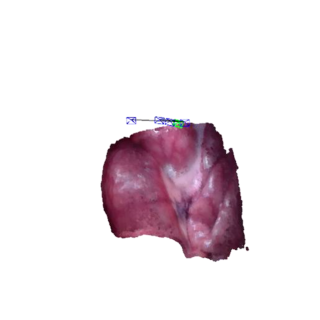}
			\end{minipage}
		}
		\subfloat[Frame 400]{
			\label{fig:scanmatching_c}
			\begin{minipage}[htpb]{0.16\textwidth}
				\centering
				\includegraphics[width=1\linewidth]{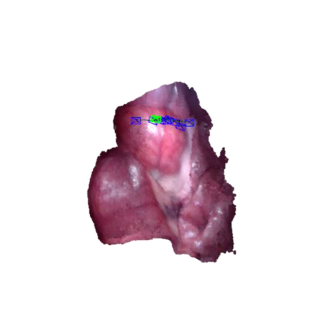}
			\end{minipage}
		}
		\subfloat[Frame 600]{
			\centering
			\label{fig:scanmatching_b}
			\begin{minipage}[htpb]{0.16\textwidth}
				\centering
				\includegraphics[width=1\linewidth]{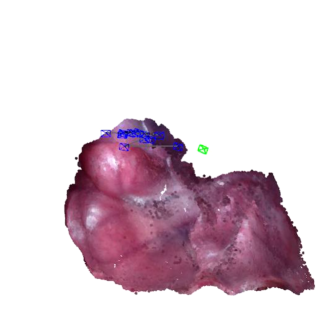}
			\end{minipage}
		}
		\subfloat[Frame 800]{
			\centering
			\label{fig:scanmatching_b}
			\begin{minipage}[htpb]{0.16\textwidth}
				\centering
				\includegraphics[width=1\linewidth]{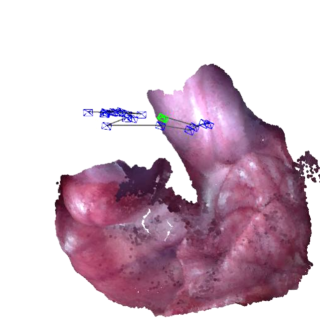}
			\end{minipage}
		}\/
		\subfloat[Frame 1000]{
			\label{fig:scanmatching_a}
			\begin{minipage}[htpb]{0.16\textwidth}
				\centering
				\includegraphics[width=1\linewidth]{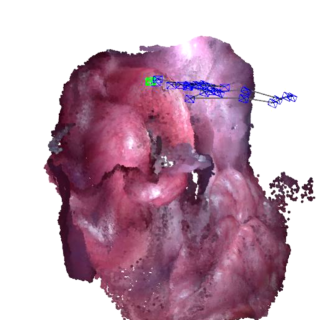}
			\end{minipage}
		}%
		\subfloat[Frame 1200]{
			\centering
			\label{fig:scanmatching_b}
			\begin{minipage}[htpb]{0.16\textwidth}
				\centering
				\includegraphics[width=1\linewidth]{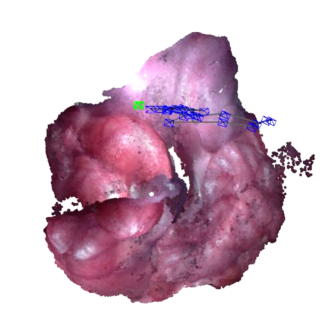}
			\end{minipage}
		}
		\subfloat[Frame 1400]{
			\label{fig:scanmatching_c}
			\begin{minipage}[htpb]{0.16\textwidth}
				\centering
				\includegraphics[width=1\linewidth]{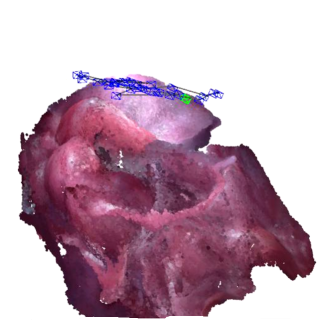}
			\end{minipage}
		}
		\subfloat[Frame 1600]{
			\centering
			\label{fig:scanmatching_b}
			\begin{minipage}[htpb]{0.16\textwidth}
				\centering
				\includegraphics[width=1\linewidth]{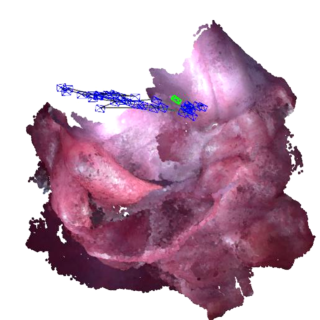}
			\end{minipage}
		}
		\subfloat[Frame 1800]{
			\centering
			\label{fig:scanmatching_b}
			\begin{minipage}[htpb]{0.16\textwidth}
				\centering
				\includegraphics[width=1\linewidth]{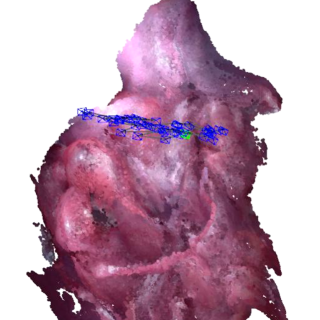}
			\end{minipage}
		}\/
		\subfloat[Frame 1]{
			\label{fig:scanmatching_a}
			\begin{minipage}[htpb]{0.16\textwidth}
				\centering
				\includegraphics[width=1\linewidth]{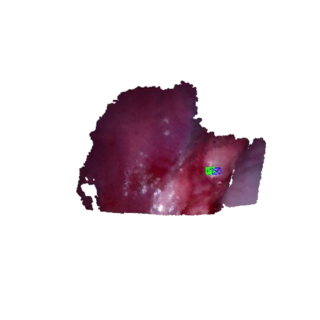}
			\end{minipage}
		}%
		\subfloat[Frame 300]{
			\centering
			\label{fig:scanmatching_b}
			\begin{minipage}[htpb]{0.16\textwidth}
				\centering
				\includegraphics[width=1\linewidth]{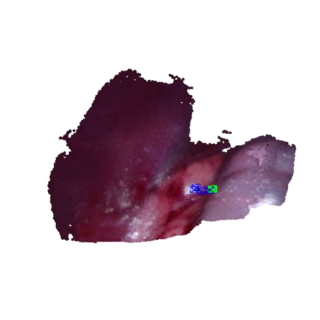}
			\end{minipage}
		}
		\subfloat[Frame 600]{
			\label{fig:scanmatching_c}
			\begin{minipage}[htpb]{0.16\textwidth}
				\centering
				\includegraphics[width=1\linewidth]{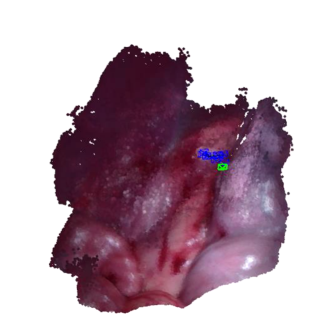}
			\end{minipage}
		}
		\subfloat[Frame 900]{
			\centering
			\label{fig:scanmatching_b}
			\begin{minipage}[htpb]{0.16\textwidth}
				\centering
				\includegraphics[width=1\linewidth]{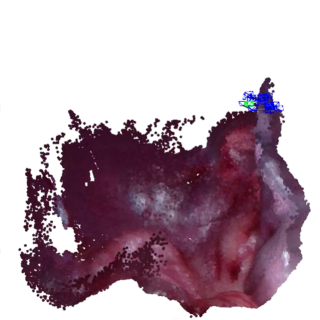}
			\end{minipage}
		}
		\subfloat[Frame 1200]{
			\centering
			\label{fig:scanmatching_b}
			\begin{minipage}[htpb]{0.16\textwidth}
				\centering
				\includegraphics[width=1\linewidth]{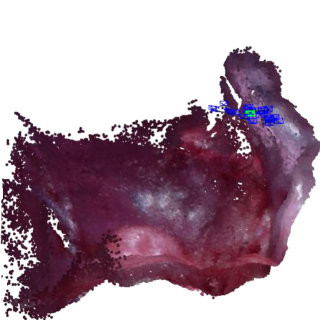}
			\end{minipage}
		}\/
		\subfloat[Frame 1500]{
			\label{fig:scanmatching_a}
			\begin{minipage}[htpb]{0.16\textwidth}
				\centering
				\includegraphics[width=1\linewidth]{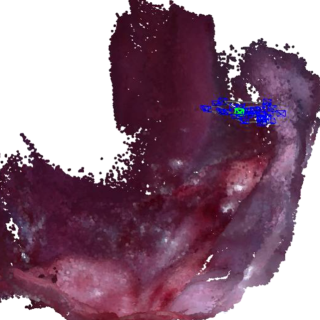}
			\end{minipage}
		}%
		\subfloat[Frame 1800]{
			\centering
			\label{fig:scanmatching_b}
			\begin{minipage}[htpb]{0.16\textwidth}
				\centering
				\includegraphics[width=1\linewidth]{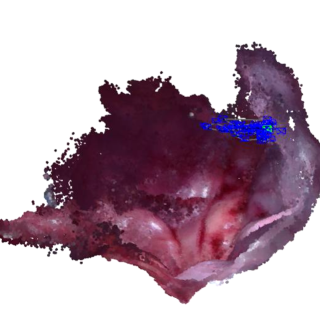}
			\end{minipage}
		}
		\subfloat[Frame 2100]{
			\label{fig:scanmatching_c}
			\begin{minipage}[htpb]{0.16\textwidth}
				\centering
				\includegraphics[width=1\linewidth]{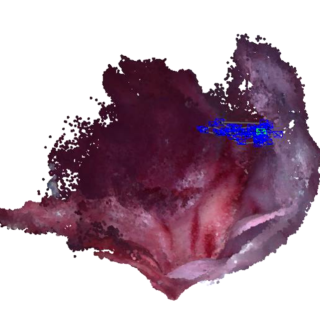}
			\end{minipage}
		}
		\subfloat[Frame 2400]{
			\centering
			\label{fig:scanmatching_b}
			\begin{minipage}[htpb]{0.16\textwidth}
				\centering
				\includegraphics[width=1\linewidth]{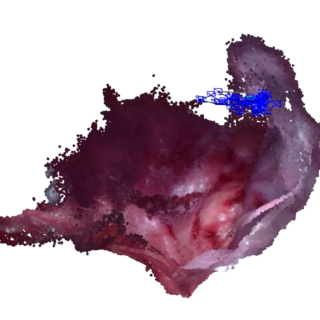}
			\end{minipage}
		}
		\subfloat[Frame 2700]{
			\centering
			\label{fig:scanmatching_b}
			\begin{minipage}[htpb]{0.16\textwidth}
				\centering
				\includegraphics[width=1\linewidth]{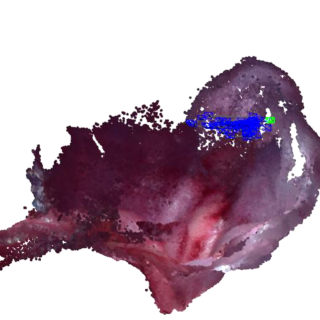}
			\end{minipage}
		}\/
		\caption{ MIS-SLAM process different soft tissues using in-vivo datasets. Pictures present the whole constructed model at different frames. The three videos are (from top to bottom): Abdomen wall, abdomen example 1 and abdomen example 2. }
		\label{fig:5model comparisons}
	\end{figure*}
	
	\section{Results and discussion}
	We validate MIS-SLAM with publicly available in-vivo stereo video datasets provided by the Hamlyn Centre for Robotic Surgery \cite{giannarou2013probabilistic}. Three videos with deformation and rigid scope movement are utilized. Other videos either have no deformation or scope motion. Please note that no extra sensing data other than stereo videos from the scope is used in the validation. The frame rate and size of in-vivo porcine dataset (model 2 in Fig. \ref{fig:5model comparisons}) is 30 frame per second and $640 \times 480$ while the other dataset is 25 frame per second and $720 \times 288$. Distance from camera to surface of soft-tissue is between 40 to 70 mm. In our previous research \cite{song2018dynamic}, due to poor quality of obtained images and some extremely fast movement of the camera, videos tested on porcine with fast or abrupt motion cannot generate good results. In this paper, however, MIS-SLAM can process large scale with much better robustness. Deformations are caused by respiration and tissue-tool interactions.  \par

	\subsection{Robustness enhancement}
	\captionsetup[subfigure]{labelformat=empty}
	\vspace{0pt} 
	\begin{figure*}[h]	
		\centering
		\subfloat[Strategy 1: Difference from ORB-SLAM]{	
			\begin{minipage}[htpb]{0.3\textwidth}	
				\centering
				\includegraphics[width=1\linewidth]{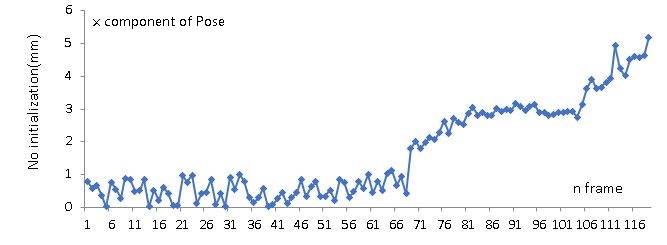}			
			\end{minipage}				
		}\/
		\subfloat[Strategy 1: Difference from ORB-SLAM]{
			\centering
			\begin{minipage}[htpb]{0.3\textwidth}
				\centering
				\includegraphics[width=1\linewidth]{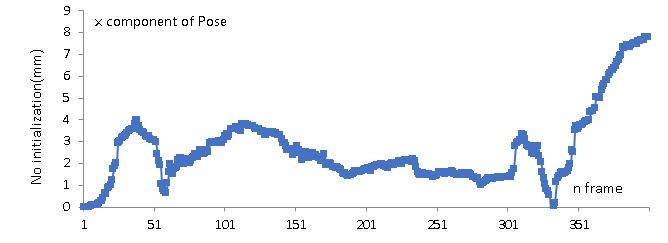}
			\end{minipage}
		}\/
		\subfloat[Strategy 1: Difference from ORB-SLAM]{
			\begin{minipage}[htpb]{0.3\textwidth}
				\centering
				\includegraphics[width=1\linewidth]{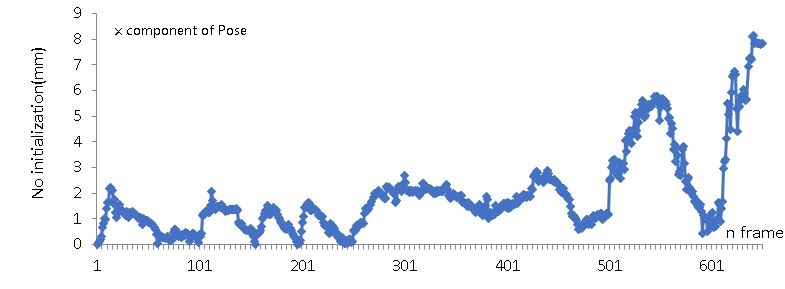}
			\end{minipage}
		}\/
		\subfloat[Strategy 2: Difference from ORB-SLAM]{
			\centering
			\begin{minipage}[htpb]{0.3\textwidth}
				\centering
				\includegraphics[width=1\linewidth]{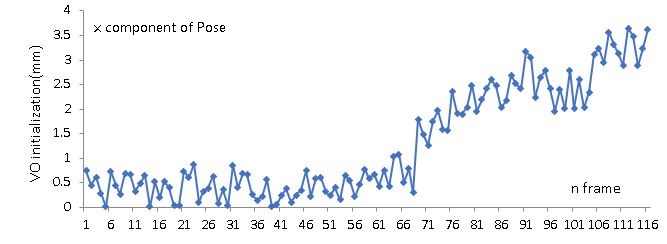}
			\end{minipage}
		}\/
		\subfloat[Strategy 2: Difference from ORB-SLAM]{
			\begin{minipage}[htpb]{0.3\textwidth}
				\centering
				\includegraphics[width=1\linewidth]{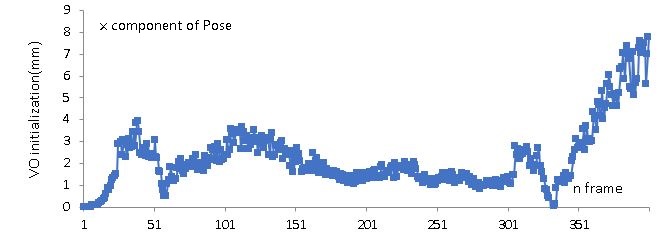}
			\end{minipage}
		}\/
		\subfloat[Strategy 2: Difference from ORB-SLAM]{
			\centering
			\begin{minipage}[htpb]{0.3\textwidth}
				\centering
				\includegraphics[width=1\linewidth]{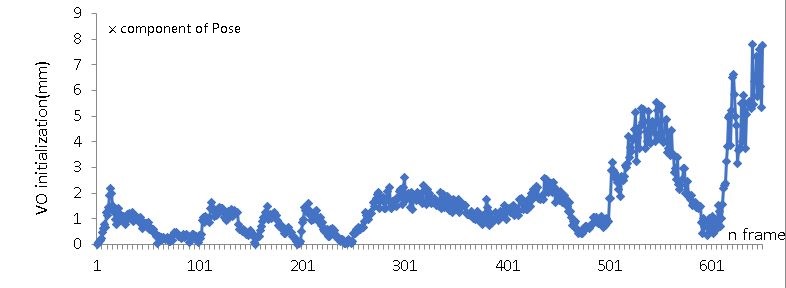}
			\end{minipage}
		}\/
		\caption{Comparisons with global pose estimation strategy includes ORB-SLAM, no initialization and conventional visual odometry. Due to no ground truth is available, we compare the distance from No initialization (Strategy 1) and visual odometry (Strategy 2) to ORB-SLAM. Results (X component of scope pose) demonstrate that both strategies lost track when cameras move fast. Our supplementary video demonstrate that ORB-SLAM still maintains the global track of the scope.}
		\label{fig:globalmodelcomparison}
	\end{figure*}
	The robustness of MIS-SLAM is significantly improved when global pose from ORB-SLAM is uploaded to GPU and employed to be an initial pose of the scope. To demonstrate the effectiveness of this strategy, we do experiments by activating and deactivating this strategy. Fig. \ref{fig:globalmodelcomparison} shows the comparisons of no global pose initialization, conventional visual odometry and initialization with ORB-SLAM pose. Due to no ground truth is available, we compare the distance between No initialization, initialized with VO and ORB-SLAM as ORB-SLAM does not fail on tracking. Graph results demonstrate that both strategies lost track when camera moves fast (errors increase abruptly). This indicates that ORB-SLAM does play an important role in deforming scenarios like computer assisted intervention when scope moves fast. Both paper and open source code from ORB-SLAM implies that author devote large efforts on enhancing localization robustness. VO is inferior to ORB-SLAM.  \par
	
	 One challenge facing reconstruction problem using stereoscope is the fast movement of scope \cite{song2018dynamic}. Configuration without global pose initialization fails to track motion when camera moves fast. Like traditional SLAM approaches, severe consequences of fast motion are the blurry images and relevant disorder of depths. These phenomena happen especially when current constructed model deforms to match the depth with false edges suffering from image blurring. Fast motion is a very challenging issue because the only data source is the blurry images. ORB-SLAM, however, is a feature based method with deliberately designed components within architectures, algorithms, thresholds, tricks and codes ensuring robustness of the system. Our supplementary videos clearly show how initialization of global pose prevents the system from failing to track camera pose.

	\begin{figure}[]		
		\captionsetup[subfigure]{labelformat=empty}
		\centering
		\subfloat[]{
			\label{fig:heart_Axial}
			\begin{minipage}[htpb]{0.4\textwidth}
				\centering
				\includegraphics[width=1\linewidth]{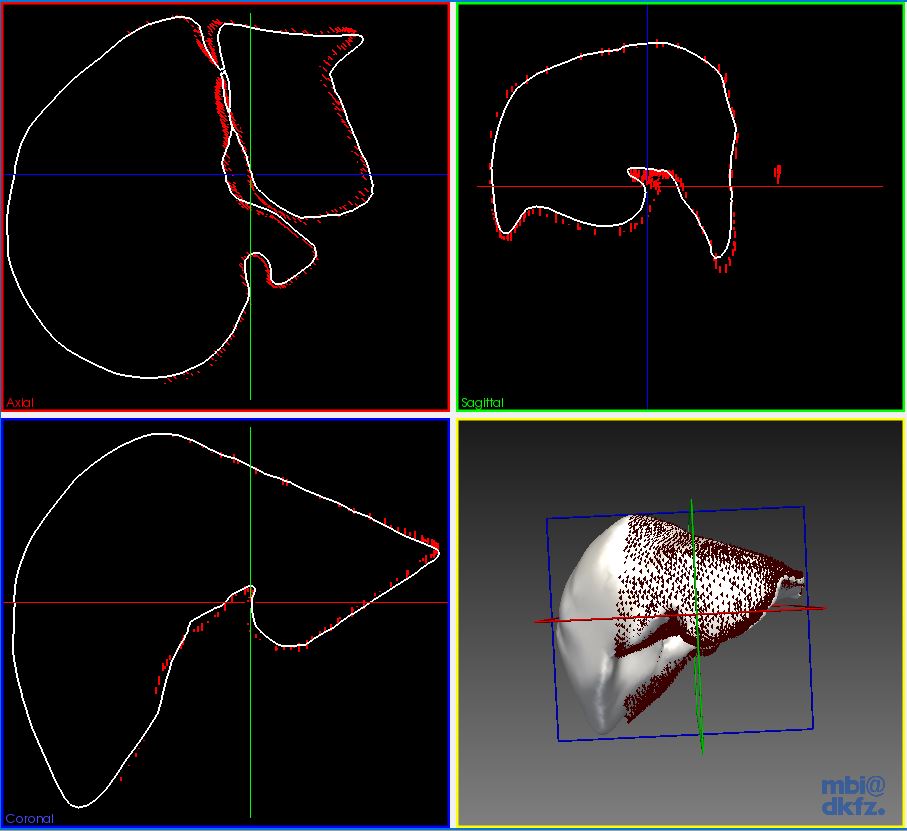}
			\end{minipage}
		}%
		\caption{The Axial, Coronal, Sagittal and 3D views of the deformed model and ground truth at the last frame (liver). The red points denote the scan of the last frame. }
		\label{fig:6traverse_map}
	\end{figure}
	\subsection{Deforming the model and fusing new depth}
	For the 6th dataset, the point cloud density is set to 0.2mm and node density is set to 4mm. For the 20/21th dataset, the point cloud density is set to 1mm and node density is set to 10mm. Point cloud downsampling process is carried out by setting a fixed box to average points fill inside each 3D box. The weights for optimization 
	are chosen as $w_{rot}=1000, w_{reg}=10000, w_{data}=1,w_{corr}=10, w_{corr}=1, w_{r}=1000000, w_{p}=1000$. A threshold is set to extract predicted visible points with point to plane distance $\epsilon_d$ as 15mm and angle threshold $\epsilon_n$ as $10^{\circ}$. We measure the error by subtracting projected model image and the observed depth image. The maximum weight is set to 20 and time stamp threshold is set to 10. Threshold of distance and angle for point to depth registration is set as 10 degree and 10mm (20mm for dataset 20/21). Truncated distance is set as 40mm (60mm for dataset 20/21).\par
	
	Threshold is employed to discard some frames when average errors are above due to low-quality depth generated from blurry images. Different from previous research, as we have good initialization of depth image, we don't suffer from lost track of camera. Fig. \ref{fig:5model comparisons} shows the results of soft-tissue reconstruction of MIS-SLAM in different frames, using 3 in-vivo laparoscope datasets \cite{giannarou2013probabilistic}. From the results it can be seen that the soft-tissues are reconstructed incrementally with texture. \par 
	The average distance of back-projection registration of the five scenarios are 0.18mm (dataset 6), 0.13mm (dataset 20) and 0.12mm (dataset 21). Dataset with ground truth still achieves 0.08mm, 0.21mm.\par
	
	\subsection{GPU implementation and computational cost}\par
	Our system is implemented on heterogeneous computing. The ORB-SLAM runs on CPU and related CPU memory. The rest system is executed on GPU. Initial global pose and ORB features are transferred to GPU for further analysis. This CPU to GPU data transferring doesn't require much bandwidth as the amount of data is fairly small. CPU initialize OpenGL for visualization framework but we utilize the interoperability from Nvidia's CUDA to directly visualize model in GPU end which saves huge amount of data transferring. Because in most cases GPU module is slower than CPU part, we utilize first-in-last-out feature in the `stack` data structure to ensure GPU always process the latest data. \par 
	
	The open source software ORB-SLAM is adopted executed on desktop PC with Intel Core i7-4770K CPU @ 3.5 GHz and 8GB RAM. We follow \cite{mahmoud2016orbslam} to tune the parameters and structures. The average tracking time is 15ms with 640x480 image resolution and 12 ms with 720x288 image resolution. As the frame rate of the three datasets are 25 or 30 fps, ORB-SLAM can achieve real-time tracking and sparse mapping.\par 
	
	By parallelizing the proposed methods for General-purpose computing on graphics processing units (GPGPU), MIS-SLAM algorithm is currently implemented in CUDA with the hardware `Nvidia GeForce GTX TITAN X'. All the process are implemented on GPU. Current processing rate for each sample dataset is around 0.07s per frame. Although MIS-SLAM does not process the feature matching, saved computation need to be spent on visualization.
	
	\subsection{Validation using simulation and ex-vivo experiments}\par
	We also validate the MIS-SLAM based on simulation and ex-vivo experiment. In simulation validation process, three different soft-tissue models (heart, liver and right kidney) are downloaded from OpenHELP \cite{kenngott2015openhelp}, which are segmented from a CT scan of a healthy, young male undergoing shock room diagnostics. The deformation of the soft-tissue is simulated by randomly exerting 2-3 mm movement on a random vertex on the model with respect to the status of the deformed model from the last frame \cite{song20163d}. We tested back-projection accuracy either not initialize global pose or do the initialization as we did in Fig. \ref{fig:globalmodelcomparison}. We randomly pick up points in the model as the accuracy is measured by averaging all the distances from the source points to target points. Fig. \ref{fig:6traverse_map} shows the final result of the simulation presented in the form of axial, coronal, sagittal and 3D maps (heart). \par
	We also tested the MIS-SLAM on two ex-vivo phantom based validation dataset from Hamlyn \cite{giannarou2013probabilistic}. As the phantom deforms periodically, we 
	do the whole process and compare it with the ground truth generated from CT scan. The average accuracies are 0.28mm and 0.35mm. \par
	\begin{table}[ht]
		\caption{Accuracy (mm) comparisons of back-projection with and without global pose initialization. Tested on three models. } 
		\centering 
		\begin{tabular}{|c|c|c|c|} 
			\hline 
			& Heart & Liver & Right kidney\\ [1.0ex] 
			\hline 
			Pose initialization & 0.41 & 0.66 & 0.62\\ 
			Without pose initialization & 0.46 & 0.68 & 0.82\\
			\hline 
		\end{tabular}
		\label{table:nonlin} 
	\end{table}

	\subsection{Limitations and discussions}\par
	One of the biggest problem in MIS-SLAM is texture blending. Results (Fig. \ref{fig:5model comparisons} and supplementary video) indicate that when camera moves, the brightness of visible region shows significant illumination differences from other invisible region. Few tissues even indicates blurry textures. The texture blending procedure involves pixel selection and blending described in Fig. \ref{Fusionstep}. If the current images are in perfect registration and identically exposed, the only significant issue involved in large scale rigid reconstruction is illuminations from different angles of light. This illumination problem cause systematic difference between two images. In MIS-SLAM, creating clean, pleasing looking texture map in our non-rigid situation is more difficult. There are many challenges in MIS-SLAM: the model has to deform according to the moving tissue, which introduce ghosting in color compositing; the camera is very close to the tissue and the exposure differs much as it moves, which resulting in visible seams in final model; image motion blurring is another problem due to the camera moves fast. We will improve color blending scheme in the future work.\par 
	
	Another improvement will be how to design a better close loop module. ORB-SLAM uses sparse features to relocate camera based on presumption that no relative motion exist in environment. In surgical vision, however, the scenario is deforming which make the case even more difficult.\par

	\section{Conclusion}
	We propose MIS-SLAM: a complete real-time large scale dense deformable SLAM system with stereoscope in Minimal Invasive Surgery based on heterogeneous computing by making full use of CPU and GPU. We significantly improved the robustness by solving unstableness caused by fast movement of scope and blurry images. Benefiting from robustness, MIS-SLAM is the first SLAM system which can achieve large scale scope localization and dense mapping in real-time. MIS-SLAM is suitable for clinical AR or VR applications when camera is moving relatively fast. Future work will be concentrated on exploring an approach to balance textures from different illumination. We will also find a way to do better close loop when previous shape is re-discovered.

	\addtolength{\textheight}{-12cm}   
	




	\bibliographystyle{ieeetr}
	\bibliography{reference}   

\end{document}